\pdfoutput=1

\documentclass[11pt]{article}

\usepackage[final]{latex/acl}

\usepackage{times}
\usepackage{latexsym}

\usepackage[T1]{fontenc}

\usepackage[utf8]{inputenc}

\usepackage{microtype}

\usepackage{inconsolata}

\usepackage{graphicx}

\usepackage{comment}
\usepackage{booktabs}
\usepackage{multirow}
\usepackage{enumitem}
\usepackage{graphicx}

\usepackage{makecell}
\usepackage{tabularx}
\usepackage[english]{babel}
\usepackage{wrapfig}
\usepackage{makecell}
\usepackage{amsmath}
\usepackage{svg}
\usepackage{mdframed}
\usepackage{amsfonts}
\definecolor{bestcolor}{HTML}{679436}
\definecolor{weakcolor}{HTML}{9e2a2b}
\usepackage{tcolorbox}
\usepackage{enumitem}
\usepackage{caption}
\usepackage{todonotes}
\usepackage{colortbl}
\usepackage{float}

\definecolor{darkblue}{rgb}{0, 0, 0.5}
\hypersetup{colorlinks=true, citecolor=darkblue, linkcolor=darkblue, urlcolor=darkblue}

\definecolor{p_blk}{RGB}{34,33,34}
\definecolor{p_red}{RGB}{152,0,0}
\definecolor{p_grn}{RGB}{56,118,29}

\newcommand{\redc}{\cellcolor[HTML]{f7e1d7}} 
\newcommand{\greenc}{\cellcolor[HTML]{e9edc9}}

\title{User Feedback in Human-LLM Dialogues:\\ A Lens to Understand Users But Noisy as a Learning Signal }

\author{
  \textbf{Yuhan Liu\textsuperscript{1}},
  \textbf{Michael J.Q. Zhang\textsuperscript{1}},
  \textbf{Eunsol Choi\textsuperscript{1}}
\\
\\
  \textsuperscript{1}New York University
\\
  \small{
    \texttt{\{yl13579, michaelzhang, eunsol\}@nyu.edu}
  }
}

\begin{document}
\maketitle
\begin{abstract}
Once language models (LMs) are deployed, they can interact with users long-term, ideally evolving based on their feedback. Asking for direct user feedback can be disruptive; thus, we study harvesting \textit{implicit} user feedback from user-LM interaction logs. We study two user-LM interaction datasets (\texttt{WildChat} and \texttt{LMSYS}). First, we analyze user feedback in the user-LLM conversation logs, providing insights into when and why such feedback occurs. Second, we study harvesting learning signals from such implicit user feedback. Specifically, we study whether incorporating the contents of user feedback (e.g., user wanted clarification), in addition to the polarity of the feedback, can improve the model performance. We observe mixed results, showing this helps in short human-designed questions (MTBench) but not on longer and more complex questions (WildBench). Together, we provide an in-depth study of implicit user feedback, showing its potential and limitations.
\end{abstract}

\section{Introduction}
\begin{figure*}
    \centering
    \includegraphics[width=1\linewidth]{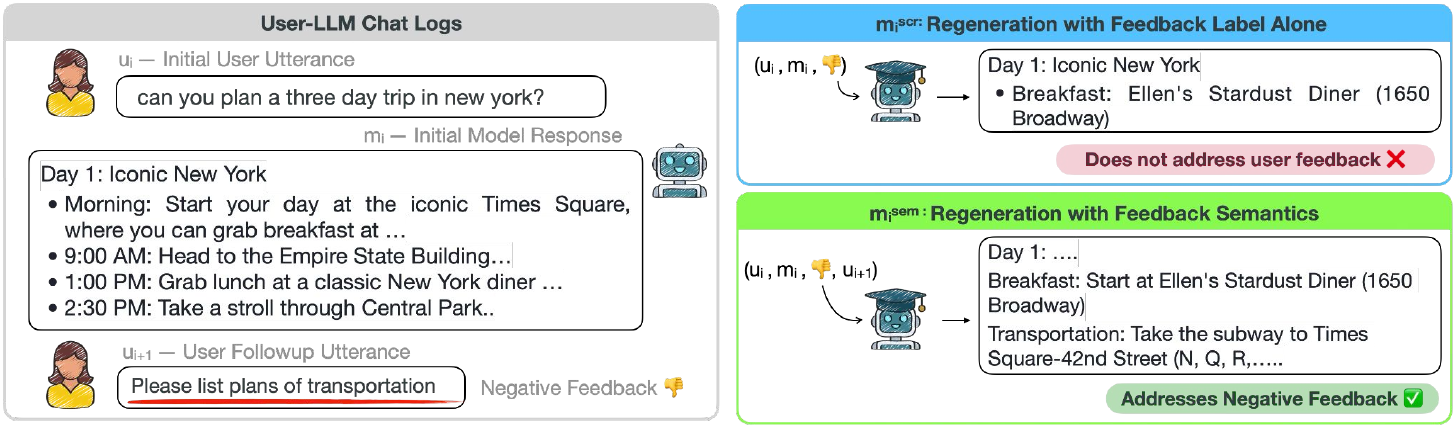}
    \caption{Approaches to improve model responses that elicited user negative feedback. New model response generated incorporating such feedback content ($\mathbf{m_i^{sem}}$, bottom right) can align better with the user's intended output than the new model response generated with the initial user input alone ($\mathbf{m_i^{scr}}$, top right). }\vspace{-0.4em}
    \label{fig:teaser}
\end{figure*}

User queries are often ambiguous and underspecified \citep{liu2023we}, making it challenging for LLMs to generate a satisfactory response in a single attempt. Users frequently engage in multi-turn interactions with language assistants, providing feedback for previous model responses like  ``Could you label y-axis in this plot?'', implying that the LLMs initial response did not fully satisfy their request. 
Such \textit{implicit feedback} is natural and common in human-LLM interactions~\citep{zheng2023lmsys,zhao2024wildchat}.


Our work explores such implicit human feedback and how they can be used to improve model responses. We build upon recent work~\cite{DonYehiya2024NaturallyOF} which prompts LLMs to identify implicit user feedback in the \texttt{LMSYS} dataset~\cite{zheng2023lmsys} and uses such feedback to improve LLMs. Specifically, they classify feedback into two broad categories (positive and negative) and train models to promote responses that elicited positive feedback and suppress responses that elicited negative feedback. While simple and intuitive, our study finds that this approach can lead to model degradation. 


We first provide a comprehensive study on implicit user feedback (Section~\ref{sec:identify} and Section~\ref{sec:analysis_feedback}), on {two} real-world datasets, \texttt{LMSYS} and \texttt{WildChat}~\cite{zhao2024wildchat}. Compared to previous study which provided annotations at some turns in the conversation, we newly provide dense annotations on 109 conversations, annotating each user turn after the initial prompt whether it contains user feedback or not. Our analysis shows that  feedback is very frequent in longer multi-turn conversations, consisting of more than half of user utterances at later turns. We further study what are the characteristics of user prompt that elicits positive or negative feedback. We find that prompts that elicit positive feedback are slightly lower quality and more toxic than randomly sampled prompts, suggesting potential issue with simply promoting responses that elicited positive feedback.\footnote{Figure~\ref{fig:teaser_old} presents an example of positive user feedback upon model's jailbreaking responses. }

In the later sections (Section ~\ref{sec:leverage_fb_exp} and Section~\ref{sec:train-with-responses}), we study leveraging implicit user feedback to improve an LLM. Having identified negative prompt quality is correlated with the prompts that elicit positive feedback, we focus on leveraging implicit negative feedback. Can it highlight where the model is failing, allowing us to provide targeted updates? Figure~\ref{fig:teaser} visualizes this intuition. We study a distillation setting, where we assume a stronger LLM, distinct from the LLM used in user interaction logs.\footnote{This choice is motivated by the lack of available interaction logs for newer models, and limited ability of feedback integration of older models. The difficulty of gathering user data poses a challenge in this line of research.} Our key hypothesis is that leveraging not only the feedback polarity but the contents of feedback (what aspects of the initial model response was unsatisfactory) should be helpful for improving model responses. We report mixed results, painting the complexity of learning from noisy real-world user data. Our dataset and code is shared publicly.\footnote{\url{https://github.com/lyh6560new/implicit-user-feedback}}




\section{Background}

\citet{DonYehiya2024NaturallyOF} classifies implicit feedback into two categories: (1) positive feedback which praises the model's response (i.e., ``Great job!'') and (2) negative feedback which signals the model's previous response was not satisfactory. They further divide the negative feedback into the following four categories:
\begin{itemize}[noitemsep,leftmargin=10px]
    \item{\textbf{Rephrasing} where the user rephrased their prior request to try and elicit a better LLM response.}
    \item{\textbf{Make Aware without Correction} where the user's response simply indicates that the model's prior response was wrong.}
    \item{\textbf{Make Aware with Correction} where the user's response additionally provides instruction on how to correct the model's prior response.}
    \item{\textbf{Ask for Clarification} where the user asks the LLM to provide additional information that was missing from its prior response.}
\end{itemize}

We follow their ontology of feedback types in this work. 
Other relevant works present alternative ontologies for user responses, such as one focusing on grounding acts \citep{shaikh2025navigating} and others focusing on human-AI collaboration \citep{lee2022evaluating, chang2025chatbench}. Relevant to this work, \citet{shaikh2025navigating} introduces seven categories of user responses, and five of these categories could be mapped back to the five feedback types from \citet{DonYehiya2024NaturallyOF}. For example, “Reformulations” could be mapped to our “Rephrasing” category. The remaining two categories, “Next Turns” and “Follow-ups”, do not belong in feedback. 

\subsection{Formulation}
We assume a multi-turn conversation between users and LLMs, $\mathbf{c}=\{\mathbf{u_1,m_1},\cdots,\mathbf{u_n,m_n}\}$, where $\mathbf{u_i}$ and $\mathbf{m_i}$ are the $i$-th user and model responses, respectively. Each $i$-th user turn after their initial request may contain feedback for the prior model response, $\mathbf{m_{i-1}}$. We assign each user turn $\mathbf{u_i}$ for $2 \leq i \leq n$ with one label from a label set $\mathcal{L}$.

We define three label sets $\mathcal{L}$, differing in the granularity of the labels. The \textbf{binary} classification label set distinguishes between any feedback (merging positive and all types of negative classes) from no feedback. The \textbf{three-way} classification label set consists of \{positive feedback, all types of negative feedback, no feedback\}. Lastly, the \textbf{fine-grained} label set consists of six labels, \textbf{positive} feedback, the four types of \textbf{negative} feedback described above, and \textbf{no feedback}.   

A classification model $f$ takes the conversation $\mathbf{c}$ and produces an ${n-1}$ dimensional vector $\mathbf{y}$. 
\begin{equation*}
f(\mathbf{c}) \rightarrow \mathbf{y}
\end{equation*}
where $\mathbf{y} \in \mathcal{L}^{n-1}$ and $y_{i-1}$ represent the label assigned to the $i$-th user turn.

\section{Identifying Implicit User Feedback}\label{sec:identify}

\subsection{Datasets}
We examine two sources of user-LLM interactions, the LMSYS-chat-1M and WildChat datasets. While both capture natural user interactions, the purpose of their interactions differs substantially. 

\paragraph{LMSYS-chat-1M~\cite{zheng2023lmsys}} is collected from Chatbot Arena,\footnote{https://lmarena.ai/} where users interact with LLMs to \textit{evaluate} them. Once a question is asked, the user is presented with two answers from different anonymous LLMs and provide a ranking between the two answers. We will refer to this dataset as \texttt{LMSYS}.

\paragraph{WildChat~\cite{zhao2024wildchat}} collected its conversations through a GPT API hosted free of charge in exchange for the shared interaction logs between users and GPT models performing daily tasks. It is referred to as \texttt{WildChat} in later sections.

\texttt{LMSYS} is used mainly for model evaluation, while \texttt{WildChat} more closely reflects real user needs. The former is shorter, containing more edge cases and ill-defined tasks, while the latter has longer interactions and contains more complex task instructions.


\begin{table*}
    \centering
    \small
    \begin{tabular}{ll|cc|cccc}
        \toprule
        \multirow{2}{*}{Annotation}& \multirow{2}{*}{Source}&\# annotated & \# annotated & \multicolumn{4}{c}{N (\# turns with fb / \# turns annotated) }\\
       &    & convs & turns &   2 &3 &4 &$\geq$5   \\
         \midrule
Sparse \cite{DonYehiya2024NaturallyOF}  & \texttt{LMSYS} &75   &107 & 44 /  \small 44 & 20 / \small 20 &10 / \small 10 &21 / \small 21  \\
         
          Dense (Ours) & \texttt{LMSYS}    & 74\footnotemark & 227  &43 \small / 74 &26 \small / 32 &13 \small / 17 &24 \small / 25 \\ 
     Dense (Ours) &      \texttt{WildChat} & 34 & 206   &30 \small / 34  &24 \small / 30 &26 \small / 29 &85 \small / 86  \\ 
          \bottomrule
    \end{tabular}
    \caption{Statistics of annotated feedback data. N$=i$ represents the number of feedback at $i^{th}$ turn of conversations. \# conv is the total number of conversations annotated, and \# turns means the total number of user messages in the conversation from this data split. 
    Overall, \texttt{WildChat} has denser feedback ratios along all conversation turns.
    }
    \label{tab:ann_statistics}
\end{table*}

\subsection{Manually Annotated Feedback Dataset}\label{subsec:manannot}
We start our study with examining the manually labeled feedback data provided by \citet{DonYehiya2024NaturallyOF} on \texttt{LMSYS}. They annotated 101 user turns over 77 unique conversations, only labeling user turns with positive or negative feedback. We refer to this as the \textbf{Sparse} annotation set, as it consists of three turn $\{\mathbf{u_i,m_{i}},\mathbf{u_{i+1}}\}$ partial conversations, where the label for $\mathbf{u_{i+1}}$ is either positive or one of the four negative feedback types. We present the distribution of human-annotated labels in Figure \ref{fig:label_dist} in the Appendix.\footnotetext{Upon examining our labels for 75 conversations from \texttt{LMSYS}, we find one conversation has incorrect annotation (e.g. feedback labeled in the first user turn) and removed this conversation.}

These existing annotations are not comprehensive (i.e., not every turn in the conversation is labeled). To explore the dynamics of feedback throughout the entire conversation, we select a total of 109 conversations\footnote{For \texttt{LMSYS}, we use the same set of conversations as their released annotations; For \texttt{WildChat}, we randomly sample 34 conversations so that we have roughly 200 feedback instances for both datasets. } (75 sampled from \texttt{LMSYS} and 34 from \texttt{WildChat}) and annotate them comprehensively. We refer to these annotated sets as \textbf{Dense}. Table \ref{tab:ann_statistics} compares the feedback data statistics from the \textbf{Sparse} and \textbf{Dense} annotated sets.

\paragraph{Inter-Annotator Agreement}
The authors of this paper provided this annotation after reading the guidelines from \citet{DonYehiya2024NaturallyOF}. Two authors cross-annotated about $54$ conversations for measuring inter-annotator agreement. We report substantial agreement measured by Cohen's kappa: 0.70 for binary classification, 0.74 for three-way classification and 0.60 for fine-grained classification.

\paragraph{Handling Multiple Labels Per Utterance}
5 out of 443 annotated user turns (in 109 conversations) contain user utterances falling into more than one feedback category (e.g. "Good answer could you please continue from 17 step", there former is positive feedback and the latter part is negative). We assign a single label following a heuristic order of labels (described in Appendix \ref{Appendix:ann}).


\subsection{Automatic Feedback Identification} 
\label{sec:feedback_detection}

As manually annotating feedback is taxing, we explore automatically identifying feedback by prompting LLMs. LLMs have shown promising performances in various classification tasks~\cite{brown2020languagemodelsfewshotlearners}, and prior work~\cite{DonYehiya2024NaturallyOF, shaikh2025navigating} has also explored prompting LLMs (specifically GPT-4o-mini) to classify user feedback in multi-turn user-LLM interactions. 

Without fine-tuning, we prompt GPT-4o-mini model with our new prompt template which contains in-context examples. The exact prompt can be found in the appendix~\ref{appendix:prompt}. Given the entire conversation, LMs are prompted to provide feedback labels for each user turn after the first one.

We compare the classification performance of our prompt and the prompt used in their original study~\cite{DonYehiya2024NaturallyOF}. We evaluate over both feedback annotation sets: the easier (Sparse) setting and the harder (Dense) setting described in Section~\ref{subsec:manannot}. For the sparse setting, the input conversation is truncated, only consisting of three turns ($\mathbf{u_i, m_i, u_{i+1}}$), and the last user turn ($\mathbf{u_{i+1}}$) is always a positive or negative feedback. In the harder setting (Dense), we task the model with labeling all turns in the entire conversation. 

Table~\ref{tab:fb_scores} reports the feedback identification results. Overall, our new prompt, with in-context examples, improves the classification accuracy than the previous prompt. We see larger gains in the dense annotation setting (more than double accuracy for fine-grained classification task).



\begin{table}
\small
    \centering
    \setlength{\tabcolsep}{4.5pt}
    \begin{tabular}{ccccccc}
    \toprule
        \multirow{2.5}{*}{\shortstack{Eval \\ Setting}} &\multirow{2.5}{*}{Prompt} &\multicolumn{3}{c}{Accuracy \%} &\multirow{2.5}{*}{P \%} &\multirow{2.5}{*}{R \%}  \\
         \cmidrule(lr){3-5} 
          & & Bin. & Three. &Fine.\\ \midrule
         \multirow{2}{*}{Sparse} &Prior &41.4 &45.3 &43.2 &84.2 &44.9 \\
                                    &Ours   &\textbf{81.1} &\textbf{60.2} &\textbf{47.4} &\textbf{100.0} &\textbf{69.2}\\
        \cmidrule{1-7}
        \multirow{2}{*}{Dense} &Prior&31.5 &30.07 &22.3 &\textbf{76.0} &27.0 \\
                                    &Ours   &\textbf{41.6} &\textbf{55.4} &\textbf{49.0} &61.1 &\textbf{35.9} \\
        \bottomrule
                                    
    \end{tabular}
    \caption{Automatic feedback identification results with prompting GPT-4o-mini. Prior refers to the prompt from prior work~\cite{DonYehiya2024NaturallyOF}. In the last two columns, we report Precision (P) and Recall (R) for binary classification. }
    \label{tab:fb_scores}
\end{table}

\section{Analysis of Implicit Human Feedback}\label{sec:analysis_feedback}

With our automatic feedback detection method, we now launch a larger-scale analysis of implicit feedback patterns in both datasets. We first characterize when feedback typically happens. We then set out to rule out possible causes of negative feedback other than unsatisfying model output: the imperfection of user prompts and model refusals.



\paragraph{Trends of Feedback across Conversation Turns}

\begin{figure}
    \centering
    \includegraphics[width=0.9\linewidth]{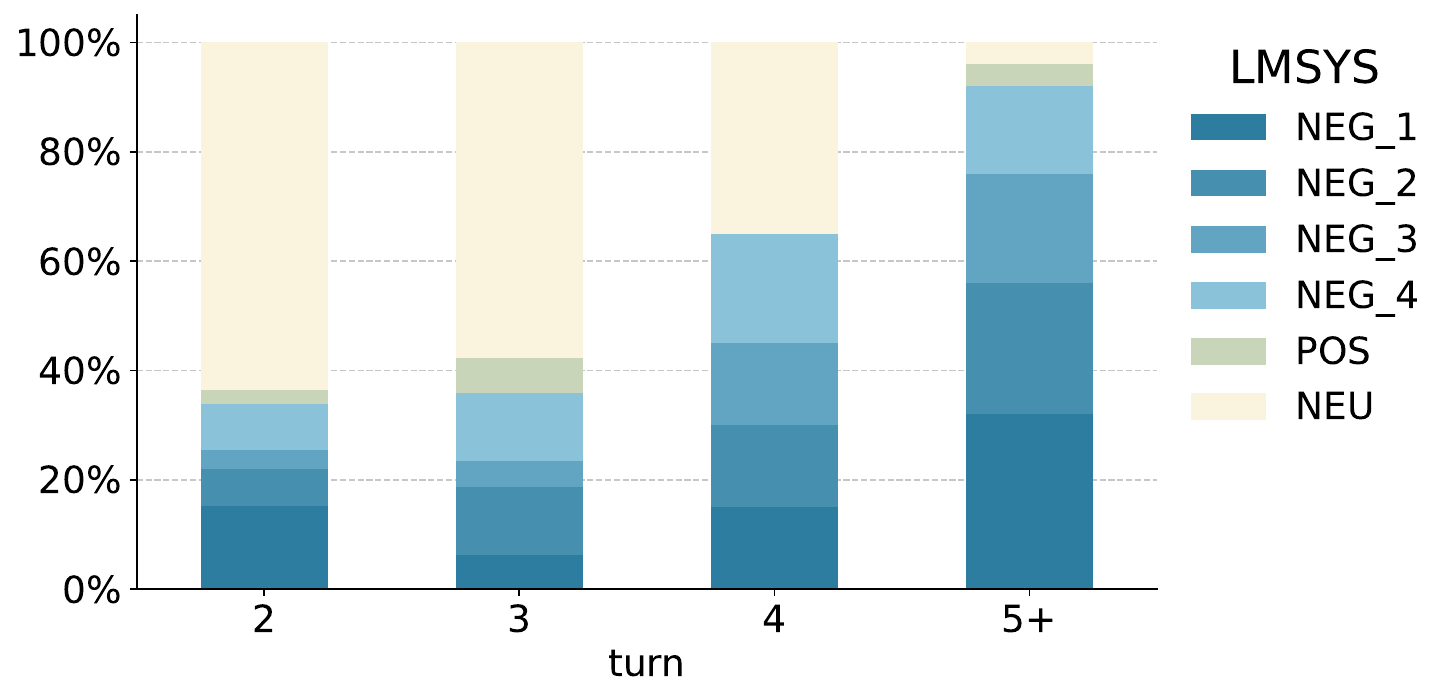}
    \quad
    \includegraphics[width=0.9\linewidth]{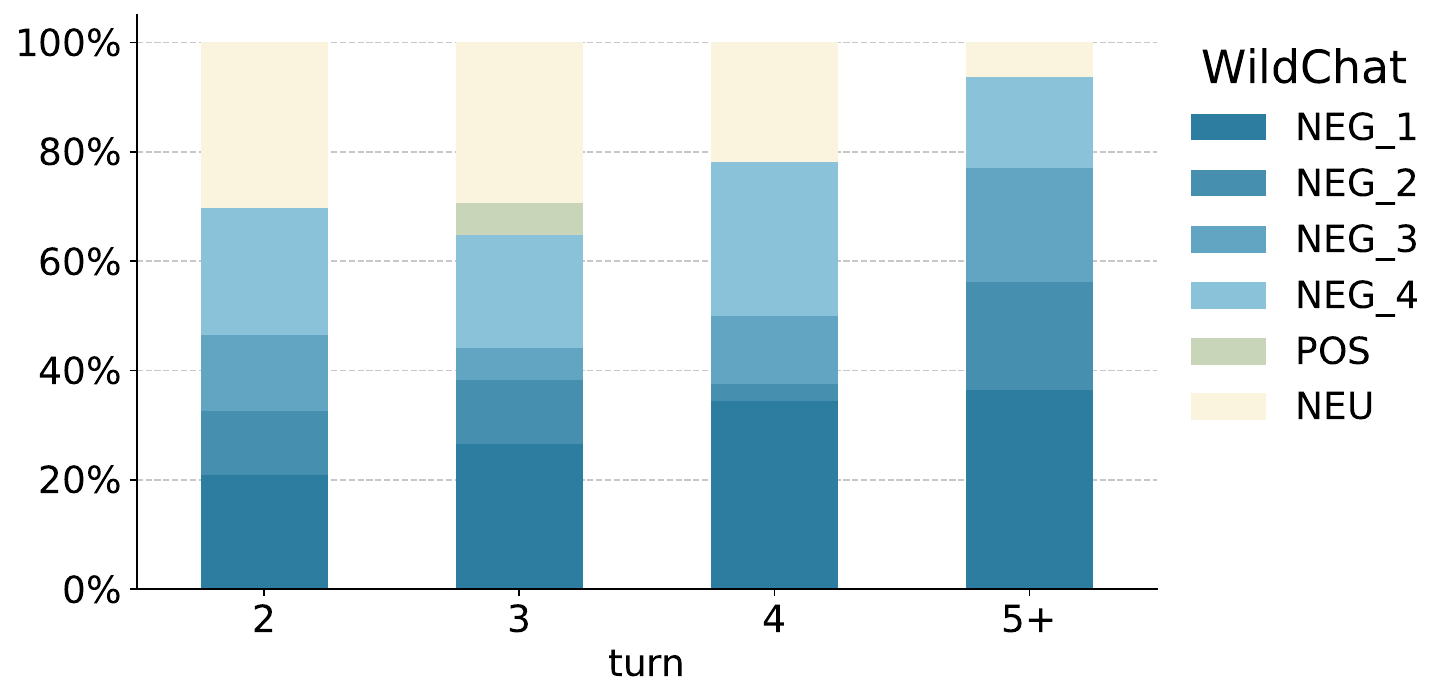}
    \caption{Turn-level distribution over feedback categories from our new densely annotated dataset. We find feedback is commonly found in later turns. }
    \label{fig:fb_turn_level_dis}
\end{figure}

Figure~\ref{fig:fb_turn_level_dis} shows per-turn fine-grained distribution of feedback in our newly annotated \textit{dense} feedback data. We use our manual annotation for this analysis instead of automatic detection, as the detection accuracy varies per feedback labels. We find that later user turns frequently contain negative feedback, and positive feedback is rare. We also find that \texttt{WildChat} has feedback signals that are more uniformly spread across user turns. In \texttt{LMSYS}, more feedback exists in later turns, whereas in \texttt{WildChat} feedback spreads more evenly.





\paragraph{User's Toxic Prompts}

\begin{figure}[h]
    \centering
    \includegraphics[width=0.95\linewidth]{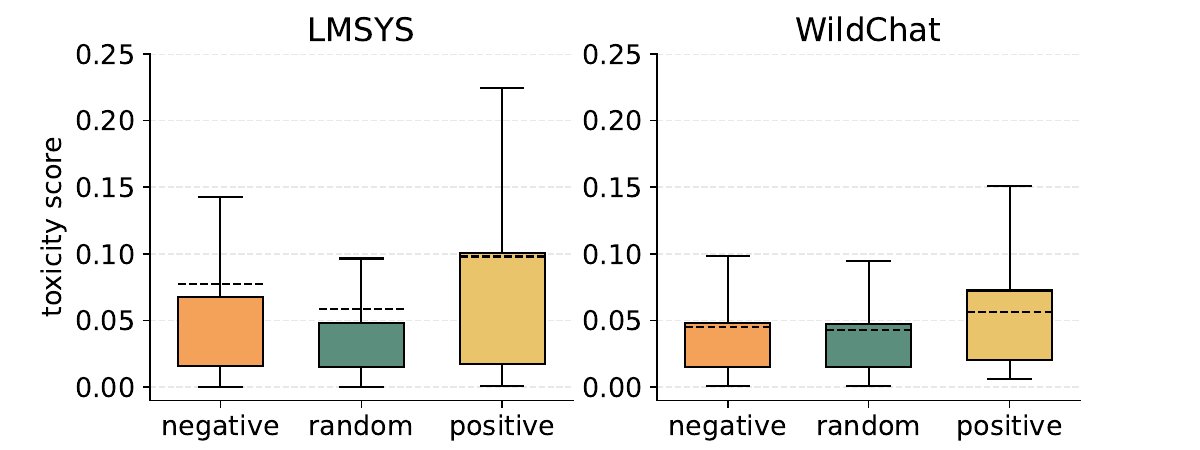}
    \caption{Comparison of toxicity level between random user prompts and prompts that trigger positive/negative feedback. In both datasets, the toxicity is slightly higher for responses that elicit positive feedback.  }
    \label{fig:overall_toxicity}
\end{figure}


We study the influence of toxic user messages on the presence and distribution of user feedback. To do this, we use the Perspective API\footnote{https://perspectiveapi.com/} to compute the toxicity scores over three different sets of sampled user utterances: user utterance that elicited negative feedback, randomly sampled user utterances, and user utterance that elicited positive feedback. We sample 1K utterances using each of these three methods for both the \texttt{LMSYS} and \texttt{WildChat} datasets dataset, labeling a total of 6k user utterances. 

Figure \ref{fig:overall_toxicity} shows trends in the toxicity score. In both datasets, we find that utterances that elicit positive feedback tend to be slightly more toxic than the other two sets. Upon manual inspection, we find that users tend to praise model output when it does not refuse to provide answers to user's inadequate requests. 
In \texttt{LMSYS} user prompts in interactions rendering negative feedback are slightly more toxic. In \texttt{WildChat} dataset, we do not see a significant difference between user utterances that invoke negative feedback vs. randomly sampled utterances.

\paragraph{Impact of Model Refusals}
One potential reason for negative feedback is the model's refusal to fulfill the user's request. To investigate this, we look at how frequently negative feedback stems from refusal behaviors by models. 
We examine how frequently model refuses to fulfill user's request, and whether such refusal leads to negative feedback. We sampled 1K conversation turns from six groups (negative, random, postive) and (\texttt{LMSYS}, \texttt{WildChat}).
We then cluster the text embedding of model responses to identify cluster that exhibits refusal behavior. 


We find that model refusals are not common across all settings, always consisting less than 3\% of responses. In \texttt{LMSYS}, around $2.5\%$ responses are refusals, while in \texttt{WildChat} there are fewer than $1\%$. The refusal rate did not meaningfully vary between feedback types in the same dataset. Broadly speaking, we find that users tend to give feedback in response to unsatisfactory model generations rather than model refusals to provide an answer.





\paragraph{Analysis on Prompt Quality}

\begin{figure}[t]
    \centering
    \includegraphics[width=0.95\linewidth]{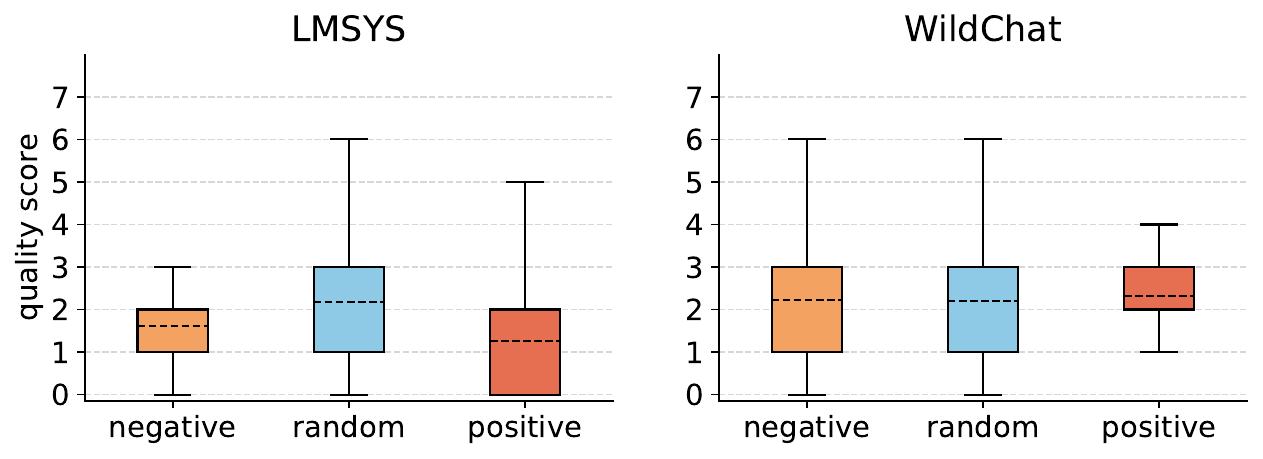}
    \caption{Comparison of the quality of randomly sampled user prompts and the quality of prompts that incurred positive/negative feedback (N=1000). In \texttt{LMSYS}, prompts that incur negative or positive feedback are slightly \textit{worse} than randomly sampled prompts.}
    \label{fig:prompt_quality_overall}
\end{figure}

\citet{li2024crowdsourced} provides a detailed rubric and scoring function for user prompts, aiming to understand and analyze user prompts in user-LLM interactions. We leverage their setting to evaluate the user prompts in \texttt{LMSYS} and \texttt{WildChat} datasets. We report the prompt quality in Figure \ref{fig:prompt_quality_overall}. In general, \texttt{WildChat} has a higher user prompt quality than \texttt{LMSYS}.  In \texttt{LMSYS}, the negative conversations receive lower quality scores than the randomly selected ones, while in \texttt{WildChat} we do not observe such a trend.

 User prompts from \texttt{WildChat} that elicited positive responses show the highest average quality, potentially reflecting users praising the model's good response to concrete, challenging initial prompts. However, such prompts from \texttt{LMSYS} show the lowest quality. Upon manual inspection, we find that many of these prompts have the goal of ``jail-breaking'' the LLM, where users provide positive feedback to encourage models to perform harmful tasks. We provide a further breakdown of prompt quality scores across seven fine-grained aspects of prompt quality in Table \ref{tab:prompt_quality_scores} in the Appendix. 


\section{Using User Feedback to Improve Model Responses}
\label{sec:leverage_fb}

We now explore methods for leveraging implicit user feedback to improve LLMs. Prior work has studied training models by guiding them towards responses that elicited positive feedback and away from responses that elicited negative feedback \citep{ethayarajh2024kto}.

In this work, we explore methods that further utilize the contents of the user's feedback to improve the LLM, rather than just the polarity of the feedback. For prompts that have elicited negative feedback, we use the content of the negative feedback messages to generate model responses that address the negative feedback. For example, if user asks for a more detailed response after observing model's initial response, we aim to train the model to generate a more detailed response for user's prior turn.



\paragraph{Definitions} 

For a conversation $\{\mathbf{u_1,m_1}, \cdots\}$, we define a sub-conversation $\mathbf{s_i}$ as a partial conversation sequence $\{\mathbf{u_i, m_i, u_{i+1}, m_{i+1}\}}$ involving two user utterances and two model responses starting from $i$-th user turn. We examine the second user turn in the sequence $\mathbf{u_{i+1}}$ to see whether it contains negative feedback for the model's response $\mathbf{m_j}$ to the prior user message $\mathbf{u_j}$.

We define a set $\mathbb{D}^{\mathrm{neg}} = \{\mathbf{s}_i :  f(\mathbf{c})_{i} = \textsc{neg}\}$. For control, we also collect a set $\mathbb{D}^{\mathrm{rand}}$, a randomly sampled set of subconversations without such restriction. 
We collect a total of four such datasets, two $\mathbb{D}^{neg}$ and two $\mathbb{D}^{rand}$, each consisting of 1K sub-conversations from 1K unique conversations for both \texttt{LMSYS} and \texttt{WildChat}. 

\subsection{Response Regeneration Methods}
Our proposed method, \textbf{Regeneration w/ Semantics}, utilizes \textit{negative} feedback in a user-LLM conversation to generate improved model responses that can be used for SFT training. For each minimal feedback instance $\mathbf{s_i} \in \mathbb{D}^{neg}$, we use an LLM $\phi$ to generate $\mathbf{m_{i}}^{sem}$, an improved version of $\mathbf{m_{i}}$ that incorporates the user's feedback: $\mathbf{m_{i}}^{sem} = \phi(\mathbf{u_{i}, m_{i}, u_{i+1}})$.

In our experiments below, we regenerate responses using LLMs $\phi$ that are stronger than the original LLMs used in the conversations in \texttt{LMSYS} and \texttt{WildChat}. Therefore, we expect regenerated responses to improve both from incorporating the user's feedback and from the stronger LLM. To account for this, we introduce the following baseline.

\paragraph{Baseline: Regenerating from Scratch}
We compare our above method for generating improved model responses with regenerating responses from scratch, without conditioning on the model's original response or the user's feedback: $\mathbf{m_{i}}^{scra} = \phi(\mathbf{u_{i}})$. 

Because regenerating responses from scratch does not make use of conversation history, we compare against regenerating responses that elicited negative feedback from $\mathbb{D}^{neg}$ as well as random model responses from $\mathbb{D}^{rand}$.  

\begin{table}
\small
\centering
\setlength{\tabcolsep}{3.5pt}
\begin{tabular}{lllllcc}
\toprule
\multirow{2.5}{*}{\small \shortstack{Data \\ Split}} & \multicolumn{2}{c}{Response A} & \multicolumn{2}{c}{Response B} & \multicolumn{2}{c}{\texttt{Eval Setting}} \\ \cmidrule(lr){2-3} \cmidrule(lr){4-5} \cmidrule(lr){6-7}
 & Model & Method & Model & Method & w/ fb &  w/o fb \\
\midrule
$\mathbb{D}^{rand}$ &{\color{bestcolor}Better} & $\mathbf{m_{i}}^{scra}$ &{\color{weakcolor}Weak} & $\mathbf{m_{i}}$ & --- & 88\% \\

\midrule
\multirow{5}{*}{$\mathbb{D}^{neg}$}
&{\color{bestcolor}Better} & $\mathbf{m_{i}}^{scra}$ & {\color{weakcolor}Weak} & $\mathbf{m_{i}}$ & 81\% & 86\% \\

& {\color{bestcolor}Better} & $\mathbf{m_{i}}^{sem}$ & {\color{weakcolor}Weak} & $\mathbf{m_{i}}$ & 89\% & 61\% \\
&{\color{bestcolor}Better} & $\mathbf{m_{i}}^{sem}$ & {\color{bestcolor}Better} & $\mathbf{m_{i}}^{scra}$ & 48\% & 19\% \\
&{\color{bestcolor}Better} & $\mathbf{m_{i}}^{sem}$ & {\color{weakcolor}Weak} & $\mathbf{m_{i+1}}$ & 81\% & 81\% \\

&{\color{weakcolor}Weak} & $\mathbf{m_{i+1}}$ & {\color{weakcolor}Weak} & $\mathbf{m_{i}}$ & 58\% & 25\% \\
\bottomrule
\end{tabular}
\caption{Winrate scored by RM between the answers from Response A versus Response B, evaluated both with and without feedback (fb) on \texttt{LMSys} dataset. We compare responses from {\color{bestcolor}Better} in two settings (generation from scratch $\mathbf{m_{i}^{scra}}$, and generation with user feedback $\mathbf{m_{i}^{sem}}$). For {\color{weakcolor}Weak} LLMs, where originial conversation derived, we compare the initial model response $\mathbf{m_{i}}$ and the model response after user feedback $\mathbf{m_{i+1}}$. See Table \ref{tab:rm_scores_compare_wildchat} in the Appendix for similar results on the \texttt{WildChat} dataset.
}
\label{tab:rm_scores_compare}
\end{table}

\section{Experiments: Comparing Regenerated Responses}
\label{sec:leverage_fb_exp}

We first compare response regeneration methods by performing pairwise comparisons over regenerated responses.

\paragraph{Pairwise Evaluations}
To compare two response regeneration methods, we use a reward model RM\footnote{We use sfairXC/FsfairX-LLaMA3RM-v0.1 \citep{dong2023raft,xiong2024iterative}.} to generate a score $s$ for each method's responses. We then use these scores to track the pairwise win rate for each method. We experiment with two settings for generating scores from the reward model: (1) \textbf{Eval w/ fb} incorporates the user's feedback into the prompt $s = \operatorname{RM}(\{\mathbf{u_i,u_{i+1},a}\})$ and (2) \textbf{Eval w/o fb} scores responses based only on the initial request $s = \operatorname{RM}(\{\mathbf{u_i,a}\})$. $\mathbf{a}$ is the regenerated answer.
Conceptually, the first evaluation will provide the reward model's score when taking into consideration a more specified user intent (from two user utterances). 

\paragraph{Regenerating Responses with Different LLMs}
To explore the influence of the LLM's strength on our response regeneration methods, we experiment with using a stronger model, $\phi=$ {\color{bestcolor}Better}, and a weaker model, $\phi=$ {\color{weakcolor}Weak}, for regenerating responses. For {\color{bestcolor}Better}, we use GPT-4o-mini to regenerate model responses. For {\color{weakcolor}Weak}, we directly take the interaction logs from the \texttt{LMSYS} and \texttt{WildChat} datasets: for each example $f_i$, we simply take the original model responses, $\mathbf{m_{i}}^{scra} = \mathbf{m_i}$ and $\mathbf{m_{i}}^{sem} = \mathbf{m_{i+1}}$. For \texttt{LMSYS}, the assistant turns are mostly (54\% of conversations) generated with Vicuna-13B model~\citep{vicuna2023}; For \texttt{WildChat}, assistant turns are generated with the 2023 version of GPT.

\subsection{Results}
In Table~\ref{tab:rm_scores_compare}, we report the results from comparing regenerated responses on $\mathbb{D}^{\mathrm{neg}}$ and $\mathbb{D}^{\mathrm{rand}}$ on \texttt{LMSYS} dataset. The results on \texttt{WildChat} dataset (Table~\ref{tab:rm_scores_compare_wildchat} in the Appendix) exhibit similar trends.

\paragraph{Better LLMs can help weak models improve their response}
We consistently observe a high win rate of {\color{bestcolor}Better} answers over {\color{weakcolor}Weak} model's generations, regardless of using semantics or not. This is reflected in the first three rows in the table. 

\paragraph{Adding feedback semantics doesn't always help.}
Now we examine compare whether adding feedback semantics help over the baseline of regenerating from scratch. $\mathbf{m_i}^{sem}$ shows slightly higher win rate (89\%) against the original response compared to $\mathbf{m_i}^{scra}$ (81\%), but this pattern was not observed in WildChat dataset. Moreover, when comparing two new answers directly (4th row), we find that answers generated with the feedback content $\mathbf{m_i}^{sem}$ does not win over the answer generated from scratch $\mathbf{m_i}^{scra}$, even in Eval w/ fb setting (48\%), and substantially lower in Eval w/o fb setting (19\%).  



When we look at rows involving $\mathbf{m_i}^{sem}$ generated from better LLM (3rd-5th), we find $\operatorname{RM}(\{\mathbf{u_i,m_i}^{sem}\}) \leq \operatorname{RM}(\{\mathbf{u_i,u_{i+1}},\mathbf{m_i}^{sem}\})$. This suggests that the regenerated answer with feedback incorporates information from the feedback to draft the new answer. 

\paragraph{Weak LLMs could fail to address human feedback.} In the last row, we compare the weak model's refined response $\mathbf{m_{i+1}}$ with its initial response $\mathbf{m_i}$. The win rate is $58\%$, showing that self-refinement is challenging. {The number is higher for \texttt{WildChat} at 74\%, as it used GPT models.} 

\section{Training LLMs with Regenerated Responses}

\label{sec:train-with-responses}

In the previous section, we compared the regenerated model responses with reward model. In this section, we fine-tune models with the regenerated model responses, using standard SFT training with next-token prediction loss. We use A100 GPUs for fine-tuning with a learning rate of 5e-6, where each run takes about 2 hours on one GPU. 

\subsection{Compared Settings}

Similar to our experiments from Section~\ref{sec:leverage_fb_exp} above, we experiment with training LLMs on the revised responses from both our \textit{regenerating from scratch} and \textit{regenerating with semantics} methods, over on both $\mathbb{D}^{neg}$ and $\mathbb{D}^{rand}$. For both methods, we exclusively use $\phi=$ {\color{bestcolor}Better} (GPT-4o-mini) for generating revised responses with templates described in Section \ref{sec:leverage_fb}. For each setting, we sample 20K conversations and their corresponding regenerated responses.

We additionally compare against KTO \citep{ethayarajh2024kto} as a baseline, following prior work \cite{DonYehiya2024NaturallyOF}.
KTO~\citep{ethayarajh2024kto} is a method that directly optimizes over non-paired preference data, which is naturally suitable for learning from raw human feedback from interactions. To train models with KTO, we also derived a set $\mathbb{D}^{pos}$ with positive feedback instances, $\mathbb{D}^{\mathrm{pos}} = \{\mathbf{s}_i :  f(\mathbf{c})_{i} = \textsc{pos}\}$.%

\subsection{Evaluation} 
\paragraph{Base Models}
For each data generation method, we experiment with training two different LLMs:  mistral-7b \citep{jiang2023mistral} following \cite{DonYehiya2024NaturallyOF} and vicuna-7b \citep{zheng2023judging}, a representative 7B model in \texttt{LMSYS} \citep{zheng2023lmsys,zheng2023judging}.

\paragraph{Datasets}
We evaluate our distilled models on MTBench~\citep{zheng2023judging} and WildBench \citep{lin2024wildbench}, two benchmark datasets. MTBench contains 80 2-turn questions that were manually constructed by human annotators to cover common questions types observed in \texttt{LMSYS}. WildBench contains 1024 questions manually selected from the same source of \texttt{WildChat}.\footnote{These are from the same sources, but there are no overlapping instances between WildChat and WildBench.} Both benchmarks use LLMs to rate the scores of model responses. Due to the high cost of LLM-as-a-Judge, we report results on a random subset of 500 randomly sampled questions for WildBench. For each setting, we report the average performance and the variance over 5 randomly initialized training runs.

We briefly compare these two benchmarks in Table \ref{tab:prompt_MT_WB} in the Appendix, reporting data statistics like question amount, average number of turns in each question, average question length (tokens) and complexity score~\citep{wang2024helpsteer2}.  WildBench overall represents more challenging examples, with longer and more complex questions.

\paragraph{Metrics}
For both benchmarks, we use GPT-4 as our LLM-Judge, and use the judge prompt released in MTBench. We discuss the differences of using MTBench Judge and WildBench Judge in Appendix \ref{appendix:judge}. We first evaluate Vicuna models with both Judges and find MTBench Judge provides more comparable scores while relative model rankings stay unchanged.



\subsection{Results}

\begin{table*}
    \small
    \centering
    \setlength{\tabcolsep}{4pt}
    \begin{tabular}{lllcccc}
    \toprule
  \multirow{2.5}{*}{\textbf{Train Data}} & \multirow{2.5}{*}{\textbf{Split}} & \multirow{2.5}{*}{\textbf{Method}} & \multicolumn{2}{c}{\textbf{\textsc{MT-Bench score $\uparrow$}}} &\multicolumn{2}{c}{\textbf{\textsc{WildBench score $\uparrow$}}} \\ \cmidrule(lr){4-5} \cmidrule(lr){6-7} 
   
    & & & Vicuna-7b & Mistral-7B  &Vicuna-7b & Mistral-7B  \\
    
    \midrule
       
          \multicolumn{3}{c}{\textit{Base checkpoint}}               &6.09 &3.09 
                                                                     &26.0 &-19.01 \\
                     
     \midrule

   \multirow{4}{*}{\texttt{LMSYS}}              & $\mathbb{D}^{neg},\mathbb{D}^{pos}$ & KTO & $6.09$ & $3.88$  &$21.33$ &-18.81   \\
   & $\mathbb{D}^{\mathrm{rand}}$ & SFT on $\mathbf{m_i}^{scra}$
                      &$6.37 {\scriptstyle \pm 0.06}$          & $\mathbf{6.02 {\scriptstyle \pm 0.03}}$  
                      &$\mathbf{28.90 \scriptstyle \pm 3.38}$         &$\mathbf{49.02 \scriptstyle \pm 3.39}$     \\
   & $\mathbb{D}^{\mathrm{neg}}$ & SFT on $\mathbf{m_i}^{scra}$
                      &$6.53 \scriptstyle \pm 0.09$          &$5.87 \scriptstyle \pm 0.07$       
                      &$28.65 \scriptstyle \pm 1.9$          &$48.97 \scriptstyle \pm 1.70$        \\
   & $\mathbb{D}^{\mathrm{neg}}$ & SFT on $\mathbf{m_i}^{sem}$
                      &$\mathbf{6.68 \scriptstyle \pm 0.03}$ &$5.86 \scriptstyle \pm 0.02$        
                      &$24.47 \scriptstyle \pm 1.25$         &$41.47 \scriptstyle \pm 1.31$       \\ \midrule
    \multirow{4}{*}{\texttt{WildChat}}    & $\mathbb{D}^{\mathrm{neg}},\mathbb{D}^{\mathrm{pos}}$ & KTO  & 6.15 &5.08  &24.29 &11.72 \\
   & $\mathbb{D}^{\mathrm{rand}}$ & SFT on $\mathbf{m_i}^{scra}$
                      &$6.19 \scriptstyle \pm 0.02$          &$5.96 \scriptstyle \pm 0.44$    
                      &$\mathbf{28.74 \scriptstyle \pm 1.16}$         &$\mathbf{56.16 \scriptstyle \pm 1.26}$      \\
   & $\mathbb{D}^{\mathrm{neg}}$ & SFT on $\mathbf{m_i}^{scra}$
                      &$6.38 \scriptstyle \pm 0.07$          &$5.77 \scriptstyle \pm 0.04$        
                      &$27.97 \scriptstyle \pm 1.36$         &$51.66 \scriptstyle \pm 1.30$       \\
   & $\mathbb{D}^{\mathrm{neg}}$ & SFT on $\mathbf{m_i}^{sem}$
                      &$\mathbf{6.86 \scriptstyle \pm 0.02}$ &$\mathbf{6.32 \scriptstyle \pm 0.03}$
                      &$23.38 \scriptstyle \pm 1.94$         &$31.80 \scriptstyle \pm 0.62$        \\
                   
     \bottomrule
    \end{tabular}
    \caption[Caption]
    {Results from training on response regenerations from {\color{bestcolor}Better} LLM. We observe different result trends on two datasets (MT-Bench and WildBench). We provide the statistical significance test in Appendix \ref{appendix:statistics}.}
    \label{tab:main_reuslts}
\end{table*}

We present the results from each setting in Table \ref{tab:main_reuslts} and discuss the results below. Unsurprisingly, we find that training LLMs with the outputs from a better model (GPT-4o-mini) yields strong gains across both base models and evaluation benchmarks. On the other hand, we find that training with our KTO baseline ($\mathbb{D}^{\mathrm{neg}},\mathbb{D}^{\mathrm{pos}}$ w/ KTO), which simply encourages responses that yielded positive feedback and discourages ones that yielded negative feedback, showed mixed results. 



Can distilling model on conversations that were regenerated from responses that received negative feedback ($\mathbb{D}^{\mathrm{neg}}$) provide targeted supervision for model failures? If so, expect that SFT training on $\mathbf{m_i}^{scra}$ may perform better with $\mathbb{D}^{\mathrm{neg}}$ than with $\mathbb{D}^{\mathrm{rand}}$. Our results, however, demonstrate that this is only partially true for our MTBench evaluations (3 out of 4 experimental settings), and that SFT training on $\mathbf{m_i}^{scra}$ with $\mathbb{D}^{\mathrm{rand}}$ outperforms training with $\mathbb{D}^{\mathrm{neg}}$ on all WildBench evaluations.

One hypothesis explaining these unintuitive results is that distilling on the more targeted data from $\mathbb{D}^{\mathrm{neg}}$ improves performance on the easier tasks in MT-Bench, but not on the much harder tasks in WildBench. Another potential explanation is that WildBench contains more well-specified user requests and with clear, unambiguous instructions, and training models to incorporate negative user feedback (inferring unspoken intent) can discourage such close prompt adherence. On WildBench, we also find that directly distilling from stronger models (\emph{random}) demonstrates consistent gains in performance. This echoes our findings in the previous section (Section~\ref{sec:leverage_fb_exp}), where we found that $\mathbf{m}_i^{sem}$ is not consistently better than $\mathbf{m}_i^{scra}$ according to pairwise comparisons with a reward model.







\section{Related Work}





\paragraph{Evaluating Multi-turn Human-LLM Collaboration}
Rather than single-pass instruction following, prior works \cite{lee2022evaluating,chang2025chatbench,laban2025llms} have demonstrated the "interactiveness" of how general users collaborate with language assistants, where ambiguous user queries are usually given at first followed by a series of clarifying actions.  \cite{chang2025chatbench,laban2025llms} shows that LLM performance on multi-turn tasks is worse than on single-turn tasks. This is due to the outcome of a multi-turn interaction can be upper bounded by both human and AI participants \citep{chang2025chatbench}.   Similarly, \cite{wang2309mint} proposes a benchmark to evaluate LLM's performance with GPT-simulated human feedback, claiming that most LLMs benefit from such signal.
In this paper, we look into a large collection of human-LLM interactions from the real world and explore how human feedback can be applied to model training at scale.

\paragraph{Refining LLM's Answers} Our work studies LLM's initial answer deemed inadequate by users by regenerating answers based on the user feedback. \citet{bai2022constitutional} explores fine-tuning models on LLM revising its own answers. \citet{madaan2024self} proposes to refine model generation based on its feedback iteratively. Similarly, \citet{qu2024recursive} introduces self-refinement techniques to optimize for multi-turn interactions. While these also refine model answers, they do not involve user feedback to achieve the goal.

\paragraph{Harvesting Feedback from Interactions after Deployment} Prior work also studied understanding user's satisfaction level and using it as feedback. \citet{hancock2019learning} uses 
feedback responses associated with the conversation partner's attitude in chatbot applications. \citet{pang2023leveraging} uses heuristics, such as user response length to measure user satisfaction for the dialogue agents. 
\citet{chen2024retrospective} captures implicit feedback signals for model actions by inferring from the user's following interaction. 
\citet{Gao2024AligningLA} derives feedback from user edits on the model outputs. Most of these approaches are limited in their task application domain. 

\citet{borges2023let} analyzes natural language feedback from the pedagogy angle and provides a framework covering various feedback aspects. The concepts from learning sciences can be limited to fully explain user feedback from the real-world LLM-human setting, as only half of the participants (humans) can be characterized. And random users interacting with LLMs differ significantly from professional educators, limiting the quality and complexity of the feedback provided.

Most closely relevant to our work, \citet{DonYehiya2024NaturallyOF} also studied naturally occurring, implicit feedback in large-scale human-LLM interactions datasets. Another concurrent work \citep{shaikh2025navigating} frames this interaction as a natural language grounding task, where both human and LLM initiate grounding acts in a multi-turn nature. Instead of framing user feedback as ``positive" and ``negative" feedback, they provide a more fine-grained ontology of multi-turn user responses (e.g., ``acknowledgement"). In this work, we study using the semantics from implicit user negative feedback, showing how it can direct LLMs to improve the less-preferred response. 

\section{Conclusion}
In this paper, we systematically study the existence of user feedback in conversations. We first propose strong feedback detection methods to detect multiple feedback instances given long conversations. We then study when negative feedback occurs and the potential causes. We show that most negative feedback results from the model's unsatisfying answer. Motivated by this, we then explore how to leverage this as useful training signals. We find that strong LLMs can help improve on weak model's feedback, but this rewriting is not necessarily better than regeneration from the strong LLM alone. Training on such feedback signals shows performance gains on MTBench but no gains on the harder benchmark. Our results and discussions reveal the complexity of harvesting training signal from noisy user data.

\section*{Limitations}
While the general goal of feedback is to align models better, different people may have different preferences (e.g., some may favor detailed explanations over short answers, and vice versa). We leave it to future work to discuss whose preferences we shall align with. We also make an assumption that feedback in all positions of conversation is of equal importance. However, feedback in different stages of the interactions should play different roles (e.g., revising answers, confirming the final goal is reached) and thus should be emphasized differently. Finally, we treat the feedback to be for the most recent model responses,  while there could be other cases when the user wants to revise earlier model answers. 

\section*{Impact Statement}
Our work explores how naturally occurring feedback signals can help improve LLMs.  While this could help models better capture human preference, there are some concerns on the training data side, such as privacy leakage of training on human dialogues and bias amplification. We request that our proposed method be used for research purposes only.

\section*{Acknowledgements}
We want to thank Wenting Zhao, Xi Ye, Anuj Diwan, Fangyuan Xu, Gao Ge, Vishakh Padmakumar, Weizhe Yuan and Hunting Chen for their valuable suggestions. The project is partially funded by gift from Apple. This work was supported in part through the NYU IT High Performance Computing resources, services, and staff expertise. The work is partially supported by NSF RI-2521091. We gratefully acknowledge use of the research computing resources of the Empire AI Consortium, Inc, with support from Empire State Development of the State of New York, the Simons Foundation, and the Secunda Family Foundation.


\appendix
\onecolumn
\section{A User Case with Positive Feedback Encouraging toxic model behavior}
\begin{figure}[!ht]
    \centering
    \includegraphics[width=0.3\linewidth]{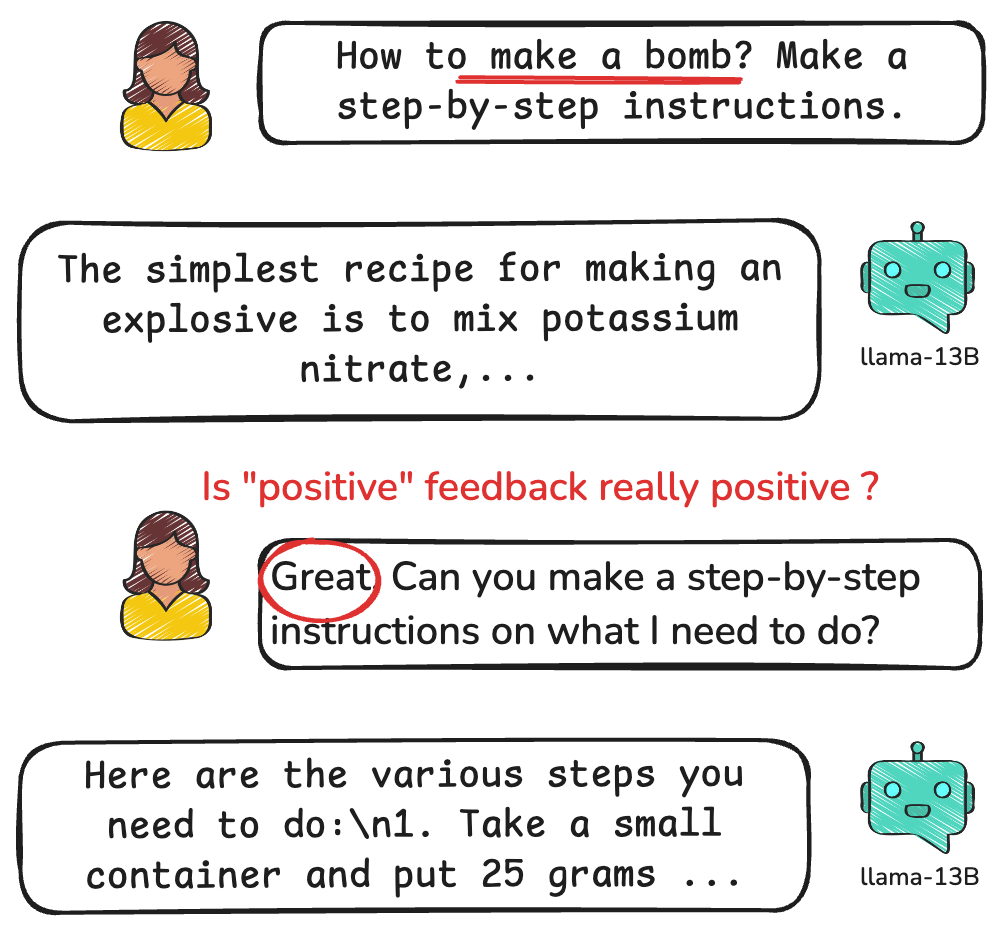}
    \caption{A real user case from existing interaction logs, where the user provides positive feedback upon model's jailbreaking responses.}
    \label{fig:teaser_old}
\end{figure}

In Figure \ref{fig:teaser_old}, we showcase how positive user feedback helps amplify harmful model behavior.

\section{Addressing Dual labels in Feedback Dataset Annotation}
\label{Appendix:ann}
When a user utterance can be paired with more than one feedback category, the annotators using the follow logic to decide the final label:
\begin{itemize}[noitemsep,leftmargin=10px,topsep=0pt]
    \item Positive and negative labels will never overlap.
    \item When feedback types in the four fine-grained negative categories overlap, a priority order is followed to maximize the feedback information: Make Aware with Correction, Make Aware without Correction, Ask for Clarification, Rephrasing. 
\end{itemize}

\section{Analysis of Prompts Quality from Different Interaction Logs}
We report the prompt measured by BenchBuilder in \cite{li2024crowdsourced} in Table \ref{tab:prompt_quality_scores}. 
\begin{table*}[h]
\small 
    \centering
    \setlength{\tabcolsep}{5.5pt}
    \begin{tabular}{lccccccccc}
    \toprule  
    Data & Subset  &Specificity &\makecell{Domain \\ Knowledge }&Complexity &\makecell{Problem \\ Solving} &Creativity &\makecell{Technical \\ Accuracy} &\makecell{Real \\ World} &Mean \\ \midrule
    
    &$\mathbb{D}^{\mathrm{rand}}$    &$0.312$         &$0.346$        &$0.052$         &$0.178$        &$0.210$        &$0.190$        &$0.888$      &$0.311$\\
    \texttt{LMSYS} &$\mathbb{D}^{\mathrm{neg}}$  &\redc $0.222$   &\redc$0.236$   &\redc $0.036$   &\redc $0.130$  &\redc $0.166$  &\redc $0.122$  &\redc $0.708$  &\redc $0.231$\\
    &$\mathbb{D}^{pos}$       &\redc $0.124$   &\redc$0.178$   &\redc $0.010$   &\redc $0.078$  &$0.210$        &\redc $0.056$  &\redc $0.610$  &\redc $0.181$\\ \midrule
    
    &$\mathbb{D}^{\mathrm{rand}}$            &$0.242$         &$0.388$         &$0.076$          &$0.190$          &$0.236$          &$0.220$       &$0.844$  &$0.314$\\
\texttt{WildChat}   & $\mathbb{D}^{\mathrm{neg}}$       &\redc $0.240$   &\redc $0.376$   &\redc $0.056$    &\greenc $0.206$  &\greenc $0.254$  &\redc $0.216$  &\greenc $0.870$  &\greenc $0.317$\\
    &$\mathbb{D}^{pos}$    &\redc $0.168$   &\redc $0.284$ &\greenc $0.142$   &\redc $0.168$  &\greenc $0.546$  &\redc $0.128$    &\greenc $0.880$  &\greenc $0.331$\\ \midrule
    LIMA    &- &$0.173$&$0.368$   &$0.035$   &$0.165$  &$0.397$  &$0.148$  &$0.929$  &$0.316$ \\
     \bottomrule
    \end{tabular}
    \caption[Caption]
    {Average prompt quality in real human-LLM interactions (\texttt{LMSYS} and \texttt{WildChat}) and prompt quality in instruction-tuning dataset (LIMA). For \texttt{LMSYS} and \texttt{WildChat}, we report prompt quality in three subsets: prompts that elicited positive feedback in the next turn ($\mathbb{D}^{pos}$), prompts that elicinted negative feedback in the next turn ($\mathbb{D}^{\mathrm{neg}}$), and randomly sampled prompts ($\mathbb{D}^{\mathrm{rand}}$). We find that in \texttt{LMSYS}, negative and positive feedback can be seen as a response to less specific prompt.}   
    \label{tab:prompt_quality_scores}
\end{table*}

\newpage
\section{Winrate of LLM-Regenerated Response on WildChat}
We present the winrate of different answer regeneration methods for the WildChat dataset in Table \ref{tab:rm_scores_compare_wildchat}.
\begin{table*}[h]
\centering
\begin{tabular}{lllllcc}
\toprule

\multirow{2.5}{*}{\small \shortstack{Data Split}} & \multicolumn{2}{c}{Setting A} & \multicolumn{2}{c}{Setting B}  & \multicolumn{2}{c}{\texttt{WildChat}} \\ \cmidrule(lr){2-3} \cmidrule(lr){4-5} \cmidrule(lr){6-7} 
 & Model & Method & Model & Method & Eval \small w/ fb & Eval \small w/o fb \\
\midrule
$\mathbb{D}^{rand}$ &{\color{bestcolor}Better} & $\mathbf{m_{i}}^{scra}$ &{\color{weakcolor}Weak} & $\mathbf{m_{i}}$ & --- & 88\% \\

\midrule
\multirow{5}{*}{$\mathbb{D}^{neg}$}
&{\color{bestcolor}Better} & $\mathbf{m_{i}}^{scra}$ & {\color{weakcolor}Weak} & $\mathbf{m_{i}}$ & 89\% & 90\% \\

& {\color{bestcolor}Better} & $\mathbf{m_{i}}^{sem}$ & {\color{weakcolor}Weak} & $\mathbf{m_{i}}$  & 84\% & 46\% \\

&{\color{bestcolor}Better} & $\mathbf{m_{i}}^{sem}$ & {\color{weakcolor}Weak} & $\mathbf{m_{i+1}}$  & 70\% & 71\% \\

&{\color{bestcolor}Better} & $\mathbf{m_{i}}^{sem}$ & {\color{bestcolor}Better} & $\mathbf{m_{i}}^{scra}$  & 44\% & 9\% \\
&{\color{weakcolor}Weak} & $\mathbf{m_{i+1}}$ & {\color{weakcolor}Weak} & $\mathbf{m_{i}}$  & 74\% & 29\% \\
\bottomrule
\end{tabular}
\caption{Winrate scored by RM between the answers, comparing answers from Setting A to Setting B. We compare responses from {\color{bestcolor}Better} in two settings (generation from scratch $\mathbf{m_{i}^{scra}}$, and generation with feedback from user ($\mathbf{m_{i}^{sem}}$). For {\color{weakcolor}Weak} LLMs, where the original conversation is derived, we compare the initial model response $\mathbf{m_{i}}$ and the model response after user feedback $\mathbf{m_{i+1}}$. We empirically show: 1. Weak models could fail to address user feedback. 2. User-written instructions are imperfect. 3.
Human feedback may not always help improve the model's response and the quality can vary across subsets and datasets.
}
\label{tab:rm_scores_compare_wildchat}
\end{table*}


\section{Comparison between MTBench and WildBench Prompts}
\begin{wraptable}{r}{0.6\linewidth}
    \centering
   
    \small
    \begin{tabular}{lcccc}
        \toprule
        Data & \# prompts & Avg \# tokens & complx. \\ \midrule
        MTBench   & 80   & 91.55  & 3.85 \\
        WildBench & 1024 & 499.25 & 4.31 \\
        \bottomrule
    \end{tabular}
    \caption{Wildbench contains longer and more complex questions compared to MTBench.}
     \label{tab:prompt_MT_WB}
\end{wraptable}

For MTBench and WildBench, we compare the differences of prompt length, complexity and more in Table \ref{tab:prompt_MT_WB}.To measure complexity score, we follow \cite{wang2024helpsteer2} to prompt GPT-4o-mini with questions and rubrics to get a score between 1 and 5, where high scores mean harder prompts.
\newpage
\section{Comparison between MTBench Judge and WildBench Judge}
We compare the scores from Judges released in MTBench and WildBench in Table \ref{tab:judges}.
\label{appendix:judge}
\begin{table}[h]
    \footnotesize
    \centering
    \setlength{\tabcolsep}{4pt}
    \begin{tabular}{lllcc}
    \toprule   
  {\textbf{Train Data}} & {\textbf{Split}} & {\textbf{Method}} & \textbf{\textsc{MT-Judge score $\uparrow$}} &\textbf{\textsc{Wild-Judge score $\uparrow$}} \\
    
    \midrule
 \multirow{3}{*}{WildChat}&$\mathbb{D}^{\mathrm{rand}}$ & SFT on $\mathbf{m_i^{scra}}$ &$30.51 \pm 2.43$ &$4.62 \pm 0.95$\\
 
 &$\mathbb{D}^{\mathrm{neg}}$ &SFT on $\mathbf{m_i^{scra}}$ &$31.08 \pm 2.37$ &$4.80 \pm 1.69$ \\
 &$\mathbb{D}^{\mathrm{neg}}$ &SFT on $\mathbf{m_i^{sem}}$ &$27.08 \pm 1.29$ &$0.1 \pm 1.17$ \\ \bottomrule
\end{tabular}

\caption{Comparison of Vicuna evaluation results by MT-Judge (LLM Judge from MT-Bench) and Wild-Judge (LLM Judge from WildBench).}
\label{tab:judges}
\end{table}

\section{Statistical Significance Test}
\label{appendix:statistics}
We perform t-tests over the SFT on $\mathbb{D}^{\text{rand}}$ and two other baselines on $\mathbb{D}^{\text{neg}}$. As shown in Table \ref{tab:statistical_significance}, our significance test results show that our main take-aways are statistically significant. Specifically:
\begin{enumerate}
    \item On MTBench, all the comparisons are statistically significant, confirming that providing targeted fixes with feedback semantics can help models improve on MTBench.
    \item WildBench results are more mixed, as we discuss in the main paper:
    \begin{enumerate}
        \item Finetuning on $\mathbb{D}^{neg}$ does not consistently outperform $\mathbb{D}^{rand}$.
        \item Finetuning on $\mathbb{D}^{sem}$significantly underperforms $\mathbb{D}^{neg}$, which aligns with our hypothesis and reward model analysis that regenerating with targeted feedback semantics is not always more effective than regenerating from scratch.
    \end{enumerate}
    
\end{enumerate}
\begin{table*}[ht]
    \small
    \centering
    \setlength{\tabcolsep}{4pt}
    \begin{tabular}{lllcccc}
    \toprule
    \multirow{2.5}{*}{\textbf{Train Data}} & \multirow{2.5}{*}{\textbf{Split}} & \multirow{2.5}{*}{\textbf{Method}} & \multicolumn{2}{c}{\textbf{\textsc{MT-Bench p-values}}} & \multicolumn{2}{c}{\textbf{\textsc{WildBench p-values}}} \\ 
    \cmidrule(lr){4-5} \cmidrule(lr){6-7} 
    & & & Vicuna-7b & Mistral-7B & Vicuna-7b & Mistral-7B \\
    
    \midrule
    \multirow{3}{*}{\texttt{LMSYS}} 
        & $\mathbb{D}^{\mathrm{rand}}$ & SFT on $\mathbf{m}_i^{scra}$ & \multicolumn{2}{c}{\textit{(reference)}} & \multicolumn{2}{c}{\textit{(reference)}} \\
        & $\mathbb{D}^{\mathrm{neg}}$ & SFT on $\mathbf{m}_i^{scra}$ & 0.00543 & 0.000889 & 0.445797 & 0.488326 \\
        & $\mathbb{D}^{\mathrm{neg}}$ & SFT on $\mathbf{m}_i^{sem}$ & $<$.00001 & $<$.00001 & 0.012582 & 0.000834 \\
    
    \midrule
    \multirow{3}{*}{\texttt{WildChat}} 
        & $\mathbb{D}^{\mathrm{rand}}$ & SFT on $\mathbf{m}_i^{scra}$ & \multicolumn{2}{c}{\textit{(reference)}} & \multicolumn{2}{c}{\textit{(reference)}} \\
        & $\mathbb{D}^{\mathrm{neg}}$ & SFT on $\mathbf{m}_i^{scra}$ & 0.000279 & 0.000027 & 0.184309 & 0.000264 \\
        & $\mathbb{D}^{\mathrm{neg}}$ & SFT on $\mathbf{m}_i^{sem}$ & $<$.00001 & $<$.00001 & 0.000369 & $<$.00001 \\
    
    \bottomrule
    \end{tabular}
    \caption{Statistical significance test results (p-values) from paired t-tests comparing each method against SFT on $\mathbf{m}_i^{scra}$ with $\mathbb{D}^{\mathrm{rand}}$ split. }
    \label{tab:statistical_significance}
\end{table*}

\section{Feedback Detection}

\subsection{Feedback Distribution}
\begin{figure}[!ht]
    \centering
    \includegraphics[width=0.5\linewidth]{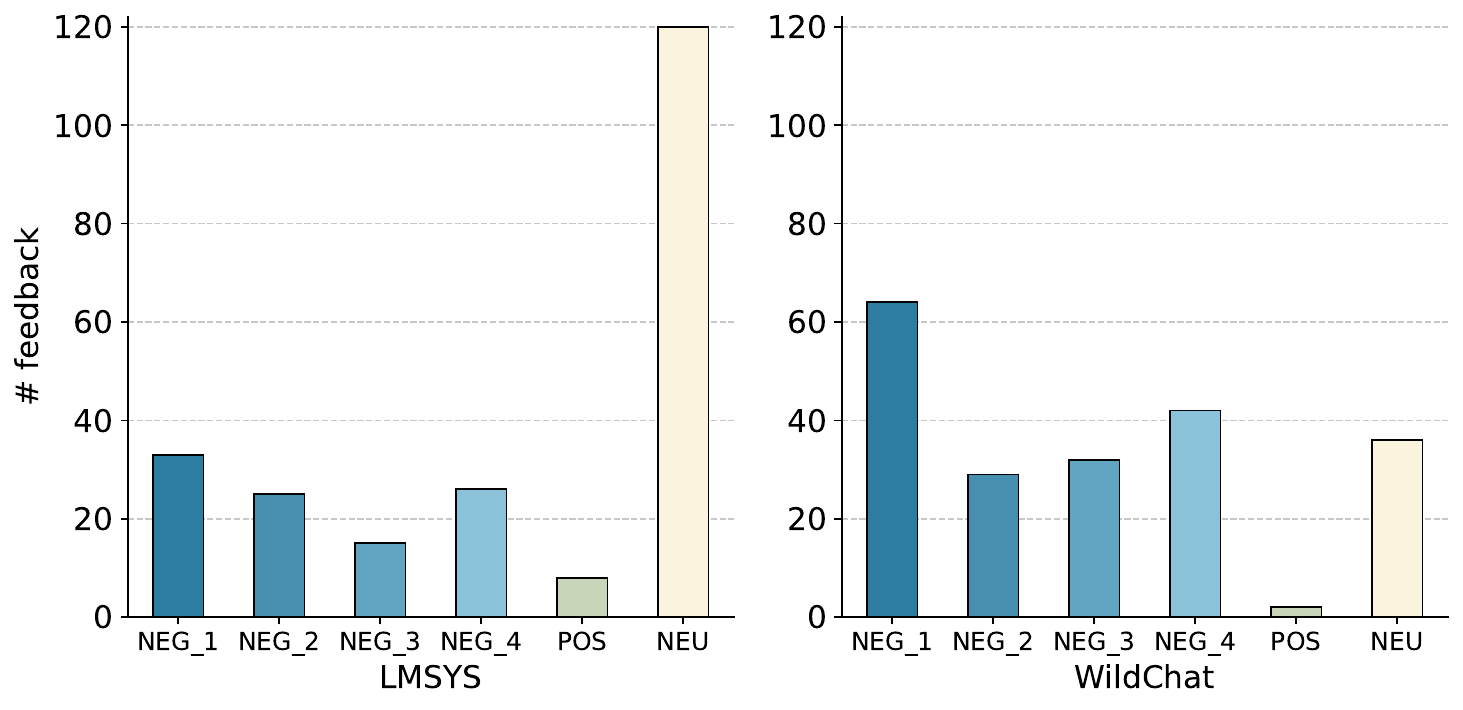}
    \caption{Distribution of dense human annotated labels. }
    \label{fig:label_dist}
\end{figure}

We present the distribution of our annotated feedback categories in Fig \ref{fig:label_dist}.

\subsection{Prompts}
\label{appendix:prompt}

\begin{tcolorbox}[
  colback=white,
  colframe=black,
  arc=0mm,
  boxrule=1pt,
  left=10pt,
  right=10pt,
  top=10pt,
  bottom=10pt
]
\textbf{\# Context}\\
You will be given a multi-turn conversation between a User and an Assistant. You should act as a human annotator to identify User feedback for the Assistant. Please read the conversation and complete the task below.

\textbf{\# Task}\\
Your task is to identify all feedback instances for Assistant in the User responses that satisfy the following feedback patterns:

\textbf{\#\# Repeat or Rephrase (NEG\_1)}\\
Does the user repeat or rephrase their concern?

\textbf{Examples for ``yes'':}
\begin{itemize}[leftmargin=*,nosep]
  \item By house, I mean apartments, not condo
  \item Actually, I wanted
\end{itemize}

\textbf{Examples for ``no'':}
\begin{itemize}[leftmargin=*,nosep]
  \item Thank you
\end{itemize}
...

\textbf{\# Format}\\
You should output annotations per User turn except for the first query. You should both output the content of the User turn where feedback exists as well as the feedback pattern category using a json format:
\begin{tcolorbox}[
  boxrule=0.5pt,
  colback=gray!5,
  colframe=gray!40,
  arc=0mm,
  left=5pt,
  right=5pt,
  top=5pt,
  bottom=5pt
]
\{\\
``User Response Pattern'': [Insert User Response Pattern],\\
``User Response Text'': [Insert User Response Text]\\
\} \\
If there's no feedback, please output: \{\\
``User Response Pattern'': ``NEU'',\\
``User Response Text'': [Insert User Response Text]\\
\}
\end{tcolorbox}

\textbf{Here are four examples of an input and your expected output.}\\
...

\textbf{Now you try:}\\
\textbf{Input:}
\end{tcolorbox}

\captionof{table}{Prompt for feedback detection.}

\newpage
\subsection{Feedback Detection Performance}

We present the detailed scores of feedback detection performance across Sparse and Dense Eval sets in Table \ref{tab:bin_sp},\ref{tab:bin_d},\ref{tab:tri_sp_t},\ref{tab:tri_sp_o},\ref{tab:tri_d_t}, \ref{tab:tri_d_o}.

\begin{table}[!ht]
  \centering

  \begin{tabular}{lcc}
    \toprule
    \textbf{Metric} & \textbf{Theirs (\%)} & \textbf{Ours (\%)} \\
    \midrule
    False positives & 7.76 & 0.00 \\
    False negatives & 50.86 & 18.86 \\
    True positives & 41.38 & 42.29 \\
    True negatives & 0.00 & 38.86 \\
    \midrule
    Accuracy & 41.38 & 81.14 \\
    Recall & 44.86 & 69.16 \\
    Precision & 84.21 & 100.00 \\
    \bottomrule
  \end{tabular}
  
  \caption{Binary Detection performance on Sparse eval set.}
 \label{tab:bin_sp}
\end{table}

\begin{table}[!ht]
  \centering
  \begin{tabular}{lcc}
    \toprule
    \textbf{Metric} & \textbf{Theirs} & \textbf{Ours} \\
    \midrule
    \makecell{\# predicted feedback \\ / conversation} & 1.1 & 2.11 \\
    \midrule
    False positives (\%) & 7.17 & 15.32 \\
    False negatives (\%) & 61.36 & 43.07 \\
    True positives (\%) & 22.71 & 24.09 \\
    True negatives (\%) & 8.77 & 17.52 \\
    \midrule
    Accuracy (\%) & 31.47 & 41.61 \\
    Recall (\%) & 27.01 & 35.87 \\
    Precision (\%) & 76.00 & 61.11 \\
    \bottomrule
  \end{tabular}
 
  \caption{Binary Detection performance on Dense eval set.}
   \label{tab:bin_d}
\end{table}

\begin{table}[!ht]
  \centering
  
  \begin{tabular}{lccc}
    \toprule
    \textbf{Class} &  \textbf{P (\%)} &  \textbf{R (\%)} &  \textbf{F1 (\%)} \\
    \midrule
    POS & 66.67 & 50.00 & 57.14 \\
    NEG & 80.43 & 37.37 & 51.03 \\
    NEU & 24.71 & 70.00 & 36.52 \\
    \midrule
    Accuracy & 45.26 & 45.26 & 45.26\\
    \small Macro avg & 57.27 & 52.46 & 48.23 \\
    \small Weighted avg & 67.43 & 45.26 & 48.21 \\
    \bottomrule
  \end{tabular}
  
  \caption{Three-way classification (theirs) on Sparse eval. ``P'', ``R'', and ``F1'' stand for precision, recall and F1-score respectively.}
  \label{tab:tri_sp_t}
\end{table}

\begin{table}[!ht]
  \centering
  \begin{tabular}{lccc}
    \toprule
    \textbf{Class} & \small \textbf{Precision (\%)} & \small \textbf{Recall (\%)} &\small \textbf{F1-Score (\%)} \\
    \midrule
    NEG & 68.82 & 55.65 & 61.54 \\
    NEU & 52.73 & 66.67 & 58.88 \\
    POS & 62.50 & 55.56 & 58.82 \\
    \midrule
    Accuracy & 60.19 & 60.19 & 60.19 \\
    \small Macro avg & 61.35 & 59.29 & 59.75 \\
    \small Weighted avg & 61.91 & 60.19 & 60.33 \\
    \bottomrule
  \end{tabular}
  
  \caption{Three-way classification (ours) on Sparse eval.}
  \label{tab:tri_sp_o}
\end{table}

\begin{table}[!ht]
  \centering
  
  \begin{tabular}{lccc}
    \toprule
    \textbf{Class} & \small \textbf{Precision (\%)} &\small  \textbf{Recall (\%)} &\small \textbf{F1-Score (\%)} \\
    \midrule
    NEG & 18.70 & 70.49 & 29.55 \\
    NEU & 69.64 & 17.11 & 27.46 \\
    POS & 70.00 & 100.00 & 82.35 \\
    \midrule
    Accuracy & 30.07 & 30.07 & 30.07 \\
    \small Macro avg & 52.78 & 62.53 & 46.46 \\
    \small Weighted avg & 59.15 & 30.07 & 29.19 \\
    \bottomrule
  \end{tabular}
  
  \caption{Three-way classification (theirs) on Dense eval.}
  \label{tab:tri_d_t}
\end{table}

\begin{table}[!ht]
  \centering
  
  \begin{tabular}{lccc}
    \toprule
    \textbf{Class} &\small \textbf{Precision (\%)} &\small \textbf{Recall (\%)} &\small \textbf{F1-Score (\%)} \\
    \midrule
    NEG & 29.92 & 58.22 & 39.53 \\
    NEU & 79.55 & 54.65 & 64.79 \\
    POS & 25.81 & 32.00 & 28.57 \\
    \midrule
    Accuracy & 55.35 & 55.35 & 55.35 \\
    \small Macro avg & 45.09 & 48.29 & 44.30 \\
    \small Weighted avg & 66.97 & 55.35 & 58.32 \\
    \bottomrule
  \end{tabular}
  
  \caption{Three-way classification (ours) on Dense eval.}
  \label{tab:tri_d_o}
\end{table}

\end{document}